\documentclass[runningheads]{llncs}
\usepackage{graphicx}
\usepackage{hyperref}

\usepackage{multirow}
\usepackage{amsmath}
\usepackage{xcolor}

\usepackage{todonotes}

\begin{document}
\title{Discriminant audio properties in deep learning based respiratory insufficiency detection in Brazilian Portuguese\thanks{Supported by FAPESP grants 2020/16543-7 and 2020/06443-5, and by Coordenação de Aperfeiçoamento de Pessoal de Nível Superior - Brasil (CAPES) - Finance Code 001. Carried out at the Center for Artificial Intelligence (C4AI-USP), supported by FAPESP grant 2019/07665-4 and by the IBM Corporation.}}
%
%
\author{Marcelo Matheus Gauy\inst{1}\orcidID{0000-0001-8902-0435} \and
Larissa Cristina Berti\inst{2} \and
Arnaldo C\^andido Jr\inst{3}\orcidID{0000-0002-5647-0891} \and
Augusto Camargo Neto\inst{1} \and
Alfredo Goldman\inst{1}\orcidID{0000-0001-5746-4154} \and
Anna Sara Shafferman Levin\inst{1} \and
Marcus Martins\inst{1} \and
Beatriz Raposo de Medeiros\inst{1}\orcidID{0000-0001-8298-0070} \and
Marcelo Queiroz\inst{1} \and
Ester Cerdeira Sabino\inst{1}\orcidID{0000-0003-2623-5126} \and
Flaviane Romani Fernandes Svartman\inst{1}\orcidID{0000-0002-9941-3934} \and
Marcelo Finger\inst{1}\orcidID{0000-0002-1391-1175}}
\authorrunning{Gauy et al.}
%
\institute{Universidade de S\~{a}o Paulo, Butanta, São Paulo - SP, Brazil
 \and
Universidade Estadual Paulista, Mar\'ilia-SP, Brazil \and
Universidade Estadual Paulista, S\~ao Jos\'e do Rio Preto-SP, Brazil
\email{marcelo.gauy@usp.br}}
\maketitle              
\begin{abstract}
This work investigates Artificial Intelligence (AI) systems that detect respiratory insufficiency (RI) by analyzing speech audios, thus treating speech as a RI biomarker. Previous works~\cite{spira2021,219270} collected RI data (\textit{P1}) from COVID-19 patients during the first phase of the pandemic and trained modern AI models, such as CNNs and Transformers, which achieved $96.5\%$ accuracy, showing the feasibility of RI detection via AI. Here, we collect RI patient data (\textit{P2}) with several causes besides COVID-19, aiming at extending AI-based RI detection. We also collected control data from hospital patients without RI. We show that the considered models, when trained on P1, do not generalize to P2, indicating that COVID-19 RI has features that may not be found in all RI types. 
\keywords{Respiratory Insufficiency  \and Transformers \and PANNs.}
\end{abstract}
\section{Introduction}

Respiratory insufficiency (RI) is a condition that often requires hospitalization, and which may have several causes, including asthma, heart diseases, lung diseases and several types of viruses, including COVID-19. 
This work is part of the SPIRA project~\cite{SPIRA-PMLD2022}, which aims to provide cheap AI tools (cellphone app) for the triage of patients by classifying their speech as RI-positive (requiring medical evaluation). Previous works~\cite{spira2021,219270} focused on COVID-19 RI. Here, we extend them to more general RI causes.

We view \textit{speech as a biomarker}, meaning that one can detect RI through speech~\cite{spira2021,219270}.
In~\cite{spira2021}, we recorded sentences from patients and a Convolutional Neural Network (CNN) was trained to achieve $87.0\%$ accuracy for RI detection. Transformers-based networks (MFCC-gram Transformers) achieved $96.5\%$ accuracy on the same test set~\cite{219270}. Here, we study multiple models in the general RI case. For that, we provide new RI data, with $26$ RI patient audios with many causes and $116$ (non-RI) control audios.
We call the data from~\cite{spira2021} P1 and the new data P2.

Transformers~\cite{219270} trained on P1 data using MFCC-grams obtain $38.8$ accuracy ($0.367$ F1-score) when tested on P2 data. Pretrained audio neural networks (PANNs)~\cite{kong2020panns} confirm this result, with CNN6, CNN10 and CNN14 trained on P1 data are comparable to~\cite{219270} on P1 test set, but achieve less than $36\%$ accuracy (less than $0.34$ F1-score) on P2 data~\footnote{Initial tests attain above $95\%$ accuracy (above $0.93$ F1-score) when training and testing on P2 data in all $4$ networks. So P2 is not harder, it is only different.}. We provide some hypotheses for this difference in Section~\ref{sec:results}.

\section{Related Work}


Transformers were proposed to deal with text~\cite{vaswani2017attention,devlin2018bert}. Later, researchers succeeded in using Transformers in computer vision~\cite{10.1145/3505244} and audio tasks~\cite{liu2020mockingjay,gong2022ssast}. Transformers benefit from two training phases: \textbf{pretraining} and \textbf{finetuning}. The former involves self-supervised training on (a lot of) unlabeled data using synthetic tasks~\cite{devlin2018bert}. The latter involves training a model extension using labeled data for the target task. One may obtain good performance after finetuning with little labeled data~\cite{devlin2018bert}. 
PANNs were proposed in~\cite{kong2020panns}. There, multiple PANNs were pretrained on AudioSet~\cite{gemmeke2017audio}, a $5000$-hour dataset of Youtube audios with $527$ classes. These pretrained models were finetuned for several tasks such as audio set tagging~\cite{kong2020panns}, speech emotion recognition~\cite{gauy2022pretrained} 
and COVID-19 detection~\cite{spirainterpretability2022}. 

\section{Methodology}

\begin{figure}[tb]
\centering
\includegraphics[width=0.7\textwidth]{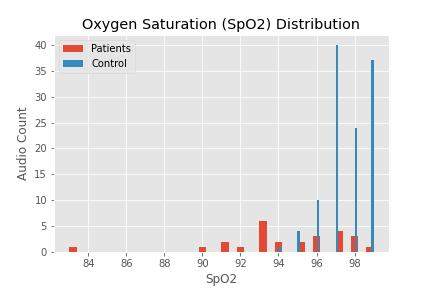}
\caption{SpO2 distribution in P2. Patient SpO2 mean  is $94.31$. For controls it is $97.66$.} \label{figure:oxygen_saturation_distribution}
\end{figure}

\paragraph{General RI dataset.}
During the pandemic, we collected patient audios in COVID-19 wards. Healthy controls were collected over the internet. This data was used for COVID-19 RI detection~\cite{spira2021,219270,gauy2023acoustic,SPIRA-PMLD2022,fernandessvartman22_speechprosody}. Now, we collect RI data with several causes from $4$ hospitals: Benefic\^{e}ncia Portuguesa (\textit{BP}), Hospital da Unimar (\textit{HU}), Santa Casa de Mar\'{i}lia (\textit{SC}) and CEES-Mar\'{i}lia (\textit{CM}). We collect three utterances: 1) a sentence\footnote{"O amor ao pr\'oximo ajuda a enfrentar essa fase com a for\c{c}a que a gente precisa"} that induces pauses, as in P1. 2) A nursery rhyme with predetermined pauses, as in P1. 3) The sustained vowel `a'. We expect the utterances to induce more pauses, occurring in unnatural places~\cite{fernandessvartman22_speechprosody}, in RI patients.
As all data was collected in similar environments, adding ward noise as in~\cite{spira2021} is no longer required and results will not be affected by bias from the collection procedure.
As a downside, controls have a health issue. Specifically, long COVID cases were not part of the $116$ controls, as we believe they could present biases~\cite{ROBOTTI2021}. Moreover, the fewer number of RI patients relative to controls (outside the pandemic) means we should use F1-score.
In P1, an RI patient was selected if his oxygen saturation level (\textbf{SpO2}) was below $92\%$. In P2, RI was diagnosed by physicians. As other factors may influence the diagnosis, RI patients often have SpO2 above $92\%$. Figure~\ref{figure:oxygen_saturation_distribution} shows SpO2 levels of patients and controls. We have $24$ RI patients and $118$ controls. However, $2$ controls had SpO2 below $92\%$. As that fits the criteria for RI, we reclassified those $2$ controls.
Lastly, we only use the first utterance as in~\cite{spira2021,219270}.
We have $14$ RI men and $12$ RI women and a mean audio duration (MAD) of $8.14s$. Also, controls comprise $36$ men and $80$ women and a MAD of $7.41s$.

\paragraph{Preprocessing.}
We break the audios in $4$ second chunks, with $1$ second steps~\cite{spira2021,gauy2023acoustic,219270}. This data augmentation prevents the audio lengths from biasing the results. 
For the MFCC-gram Transformers, the audios are resampled at $16kHz$\footnote{Performance difference by resampling the audios is minimal.}. We extract $128$ MFCCs as in~\cite{219270,gauy2023acoustic}.
For the PANNs, we do the processing steps from~\cite{kong2020panns}.

\section{Results and Discussion}
\label{sec:results}

Table~\ref{table:first_second_phase_results} shows that P2 is substantially different from P1.
We take the pretrained MFCC-gram Transformers from~\cite{gauy2023acoustic}, and fine-tune it $5$ times, with a learning rate of $10^{-4}$, batch size $16$ and $20$ epochs, on P1 training set of~\cite{spira2021}, to obtain models with above $95\%$ accuracy on P1 test set of~\cite{spira2021}. The best model on P1 validation set of~\cite{spira2021} after each epoch is saved. These $5$ models attain an average accuracy of only $38.8\%$ on P2. Additionally, we do the same with CNN6, CNN10 and CNN14. We take them from~\cite{kong2020panns}, and fine-tune~\footnote{Again, we use $20$ epochs, batch size $16$, learning rate $10^{-4}$ and best models are saved.} them $5$ times each, thus obtaining models with above $95\%$ accuracy on P1. These $5$ models of the $3$ CNNs attain an average of less than $36\%$ accuracy on P2.

\begin{table}
\caption{Performance on P2, after training on P1 training set.}\label{table:first_second_phase_results}
\begin{tabular}{|p{4.5cm} | p{3cm}| p{3cm}|}
\hline
Model & P2 F1-score  & P2 Accuracy \\ \hline

CNN6  & $0.3243\pm 0.052$           & $32.67\pm 5.34$ \\ \hline
CNN10 & $0.3226\pm 0.019$  & $33.56\pm 2.20$      \\ \hline
CNN14 &  $0.3371\pm 0.044$          & $35.39\pm 5.35$      \\ \hline
MFCC-gram Transformers &  $0.3674\pm 0.037$          & $38.82\pm 4.93$     \\ \hline

\end{tabular}
\end{table}

\begin{figure}[htb]
\includegraphics[width=0.5\textwidth]{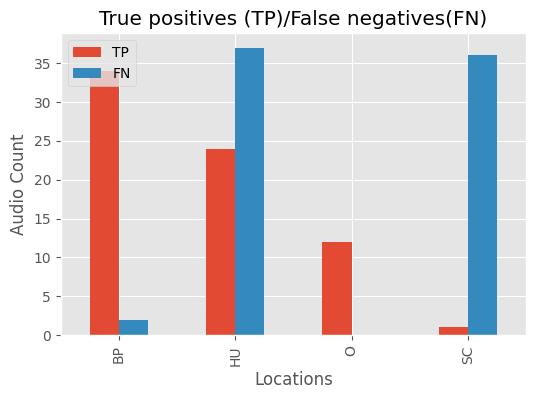}
\includegraphics[width=0.5\textwidth]{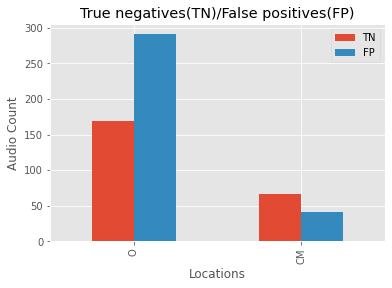}
\caption{RI patient audio count according to the hospital the data was collected.}\label{figure:tp_fn_distribution}
\end{figure}

Figure~\ref{figure:tp_fn_distribution} shows the error distribution on P2 for MFCC-gram Transformers according to the hospital~\footnote{`O' (Other) and `CM' represent controls. The other hospitals refer only to patients.} the data was collected~\footnote{Other angles do not add much. Using the PANNs yields similar results.}.
The left side shows true positives (\textit{TP}) and false negatives (\textit{FN}) of P2 RI patients. Almost all from `BP' and the $2$ `O' files (not diagnosed with RI but low SpO2) were TP. Most of the `HU' as well as almost all from `SC' were FN. We can see two reasons for the discrepancy: 1) COVID-19 patients are more numerous in `BP' than `HU' or `SC'; 2) RI patients from `BP' are more severe cases than `HU' or `SC'. As P1 was collected during the pandemic, it is filled with severe cases. 
The right side shows true negatives (\textit{TN}) and false positives (\textit{FP}) of P2 controls. `CM' were mostly TN while `O' were mostly FP. It is possible that certain comorbidities (more common in `O' than `CM') led the model to errors as it only trained on severe RI patients contrasted with healthy controls.

Thus, our results suggest that it is possible to identify the RI cause via AI, as different forms of RI have distinct audio features that are learned by the models. However, this task will require considerably more data on each RI cause.

\section{Conclusion and Future work}

We presented new RI data expanding on P1~\cite{spira2021}. RI in P2 data has many causes such as asthma, heart diseases, lung diseases, unlike P1 (COVID-19 only). Our results suggest P1 and P2 have relevant differences as AI models trained on P1 data perform poorly on P2 data. Thus some audio properties of COVID-19 RI are distinct from general RI causes, which should be identifiable.

Future work involves the expansion of P2 data so we may train models that detect RI as well as its cause. This would benefit more complex models as currently CNN6 and CNN10 outperform CNN14 and MFCC-gram Transformers.

%
%
%
\bibliographystyle{splncs04}
\bibliography{sample-ceur}
\end{document}